\documentclass[11pt,letterpaper]{article}
\usepackage{cogsys}
\usepackage[T1]{fontenc}
\usepackage{times}
\usepackage[pdftex]{graphicx} % use this when importing PDF files
\usepackage{subcaption} % Subfigures

\usepackage{natbib}
\usepackage{amsmath}
\usepackage{amsfonts}
\usepackage{amssymb}
\cogsysheading{}{2025}{}{08/2025}{10/2025}
\ShortHeadings{Robustness to Forgetting via Concept Formation}
              {N.\ Barari, E.\ Kim, C.\ J.\ MacLellan}

\begin{document} 

\title{Explaining Robustness to Catastrophic Forgetting Through Incremental Concept Formation}
 
\author{Nicki Barari}{nicki.barari@drexel.edu}
\address{Drexel University, 
         Philadelphia, PA 19104 USA}
\author{Edward Kim}{ek826@drexel.edu}
\address{Drexel University, 
         Philadelphia, PA 19104 USA}
\author{Christopher MacLellan}{cmaclell@gatech.edu}
\address{Georgia Institute of Technology, Atlanta, GA 30332 USA}

\vskip 0.2in
 
\begin{abstract}
Catastrophic forgetting remains a central challenge in continual learning, where models are required to integrate new knowledge over time without losing what they have previously learned. In prior work, we introduced Cobweb/4V, a hierarchical concept formation model that exhibited robustness to catastrophic forgetting in visual domains. Motivated by this robustness, we examine three hypotheses regarding the factors that contribute to such stability: (1) adaptive structural reorganization enhances knowledge retention, (2) sparse and selective updates reduce interference, and (3) information-theoretic learning based on sufficiency statistics provides advantages over gradient-based backpropagation.
To test these hypotheses, we compare Cobweb/4V with neural baselines, including CobwebNN, a neural implementation of the Cobweb framework introduced in this work. Experiments on datasets of varying complexity (MNIST, Fashion-MNIST, MedMNIST, and CIFAR10) show that adaptive restructuring enhances learning plasticity, sparse updates help mitigate interference, and the information-theoretic learning process preserves prior knowledge without revisiting past data. Together, these findings provide insight into mechanisms that can mitigate catastrophic forgetting and highlight the potential of concept-based, information-theoretic approaches for building stable and adaptive continual learning systems.
\end{abstract}

\section{Introduction} 
Recent advances in computer vision have been driven largely by deep neural networks, which now match or even surpass human performance on tasks such as image classification and object detection \citep{he2016deep}. Despite these successes, neural networks remain limited in their ability to learn continuously. When trained on tasks sequentially, they often suffer from \textit{catastrophic forgetting} \citep{mccloskey1989catastrophic}, a phenomenon in which newly acquired knowledge overwrites previously learned information. To mitigate forgetting, networks typically require access to previous training data or surrogate memory mechanisms, which increases computational cost and highlights a persistent trade-off between efficiency and stability.

In contrast, human learning is characterized by the gradual accumulation of knowledge across domains, where new skills and concepts are integrated with minimal disruption to prior knowledge \citep{barnett2002and, calvert2004handbook}. While humans do forget, this process is gradual and rarely catastrophic. Achieving similar stability in artificial systems is a central goal of continual learning research, which has produced methods ranging from replay buffers and parameter regularization to dynamic network architectures \citep{french1999catastrophic, wang2023comprehensive}. These 
approaches have advanced the field but face challenges, such as high memory requirements, task-specific tuning, and limited scalability as the number of tasks grows.

In prior work, a novel framework called \textit{Cobweb/4V} was introduced \citep{barari2024incremental}, extending the psychologically inspired Cobweb family of concept formation models 
\citep{fisher2014concept, gennari1989models}. Without relying on replay buffers, regularization constraints, or parameter isolation, Cobweb/4V demonstrated empirical robustness to catastrophic forgetting in continual visual learning tasks. These findings suggested that its resilience may arise from fundamental differences in how it learns and organizes knowledge.

This paper builds on those results by investigating the mechanisms that enable Cobweb/4V to retain prior knowledge while learning new concepts. Three hypotheses are considered: first, that adaptive restructuring of the concept hierarchy contributes to stability; second, that sparse, selective updates reduce interference; and third, that the information-theoretic learning mechanism plays a central role in preventing forgetting. Through a series of controlled experiments across multiple datasets, we aim to disentangle these factors and provide a deeper understanding of why Cobweb/4V resists catastrophic forgetting.

\section{Background}
\subsection{Continual Learning and Catastrophic Forgetting}

Continual learning refers to the ability of a model to incrementally acquire new knowledge over time without erasing or overwriting previously learned information. This capability is central to building adaptive, generalizable, and data-efficient systems that more closely resemble human learning \citep{mccloskey1989catastrophic}.
In artificial systems, continual learning offers key advantages: it improves adaptability to changing environments, reduces the need for retraining from scratch, and supports more efficient use of data over time.
% S
These benefits make it especially valuable across real-world applications and dynamic modeling that must respond to evolving data streams and maintain operational efficiency \citep{barari2025continual, izadkhah2024enhancing, izadkhah2025time}.

Despite its promise, continual learning remains a difficult challenge due to \textit{catastrophic forgetting}, a phenomenon where learning new information interferes with or overwrites previously acquired knowledge. This issue was first identified in early studies on sequential learning in neural networks \citep{mccloskey1989catastrophic}, and remains a central obstacle in developing flexible, adaptive learning systems.
Catastrophic forgetting happens due to a trade-off between plasticity and stability. A model must be plastic enough to adapt to new data, yet stable enough to preserve what it has already learned. Many learning systems, especially those with fixed capacity, struggle to maintain this balance. As new knowledge is encoded, previously learned information may be lost due to interference.
The issue of new knowledge overwriting prior patterns arises in many applied domains, such as those that must track evolving user interests or dynamic group preferences \citep{izadkhah2023deep, izadkhah2024enhanced}, in ways analogous to catastrophic forgetting.
% A key factor underlying this problem is the limited capacity of learning models. Systems with finite representational resources often struggle to integrate new knowledge without disrupting old memories. 
This challenge is not unique to neural networks; it arises in any learning system that must operate under capacity limits. Research in both biological and artificial domains has explored ways to mitigate this constraint, such as using hierarchical or distributed representations to reduce interference and preserve previously acquired knowledge \citep{french1999catastrophic, parisi2019continual}. Another possible contributor to forgetting in neural networks is the use of backpropagation, which adjusts all parameters of the model during training. While effective for performance on isolated tasks, these global updates tend to overwrite weights associated with earlier data, increasing the risk of interference \citep{goodfellow2013empirical}. In contrast, biological systems tend to rely on more localized updates. Mechanisms such as Hebbian learning and other neuro-inspired approaches emphasize gradual and selective modification of memory traces, offering a potential path forward for continual learning models \citep{miconi2018differentiable}. Sparsity is another factor that help the system to retain knowledge. In the brain, only a small subset of neurons activates in response to a given stimulus, which helps limit overlap between representations and preserve past knowledge. Sparse activation has been shown to support memory retention by minimizing interference between tasks \citep{olshausen1996emergence, barari2021linking}. Similarly, artificial models that use sparse representations, activating only a few components per input, tend to exhibit greater resilience to forgetting \citep{masse2018alleviating}.

Over the years, a variety of strategies have been proposed to mitigate catastrophic forgetting, ranging from memory-based replay and regularization techniques to dynamic network expansion and sparsity constraints. Although most of these approaches have been developed in the context of deep learning, the challenge extends to a broader class of machine learning systems, including those designed to emulate human-like cognitive processes. Advancing our understanding of the limitations in structural capacity, learning strategies, and representational efficiency may therefore provide critical insights for building more robust and cognitively inspired models of continual learning.

In light of this, we previously introduced \textit{Cobweb/4V} \citep{barari2024incremental}, a hierarchical concept formation approach inspired by psychological theories of human learning, that differs fundamentally from conventional deep learning approaches. Without relying on replay buffers, regularization constraints, or parameter isolation, Cobweb/4V demonstrated strong empirical robustness to catastrophic forgetting in continual visual learning tasks. To contextualize its performance, we benchmarked Cobweb/4V against a replay-based baseline, one of the most widely adopted strategies for mitigating forgetting, in order to assess its comparative effectiveness.

In the following sections, we provide a brief overview of the Cobweb framework and the  key modifications introduced in Cobweb/4V that enable it to operate on high-dimensional image data. We then turn to the central goal of this work: identifying mechanisms that contribute to resilience to forgetting.

\subsection{Cobweb - A Hierarchical Concept Formation Approach for Continual Learning} 
The Cobweb framework offers incremental and unsupervised learning from a continuous stream of examples \citep{fisher1987knowledge,fisher2014concept}. It processes incoming instances, and builds a hierarchical structure of concepts, drawing inspiration from human concept formation. Each instance is described as a collection of discrete attribute–value pairs, for example {\{color: blue; shape: square\}}. In the resulting Cobweb tree, each concept node maintains a table of attribute value probabilities that summarize the instances it has adopted.

\begin{figure}[t!]
  \centering
  \begin{subfigure}[b]{0.73\textwidth}
    \includegraphics[width=\textwidth]{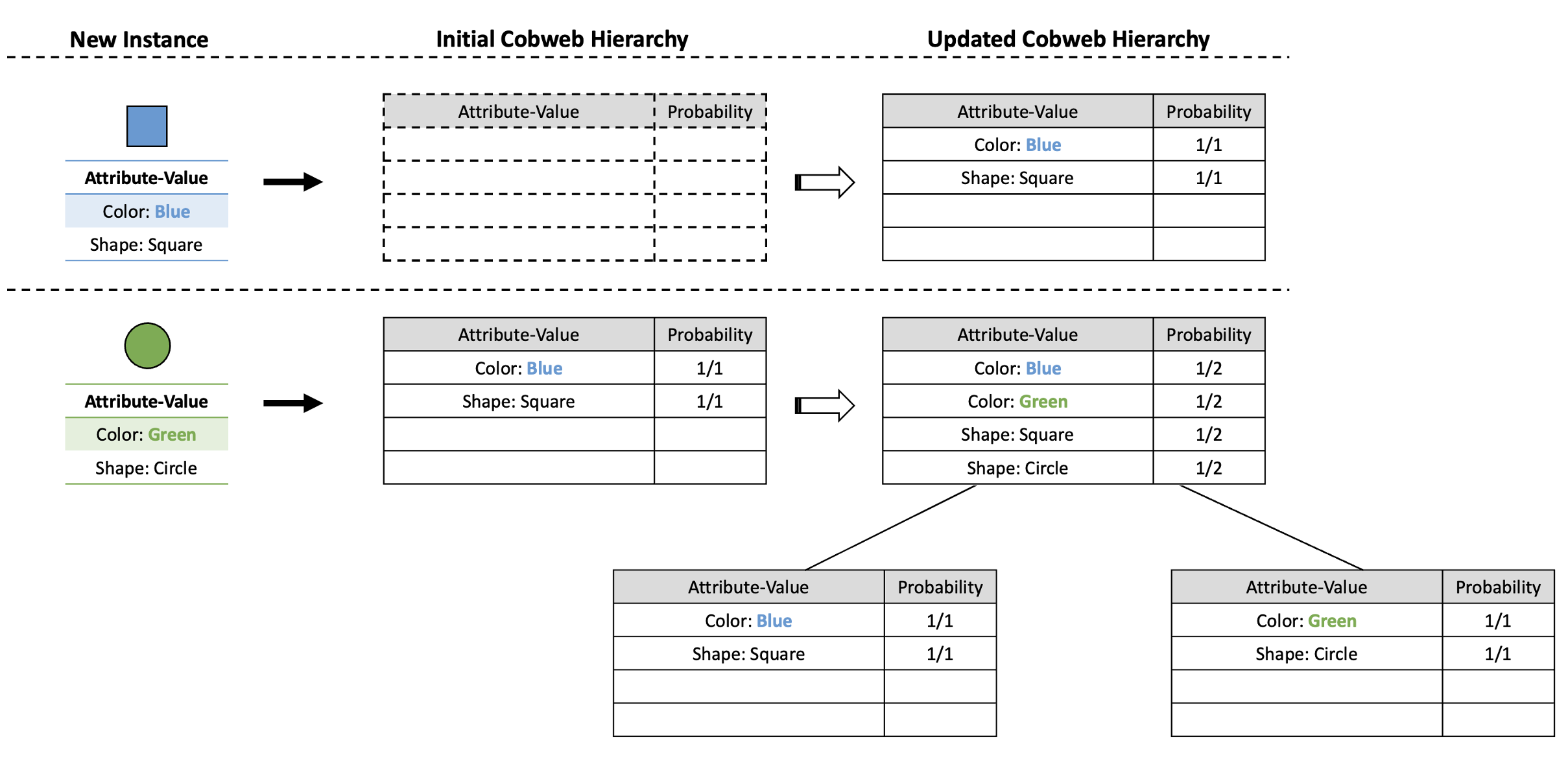}
    \label{fig:cobweb_learning}
    \caption{}
  \end{subfigure}
  \hfill
  \begin{subfigure}[b]{0.55\textwidth}
    \includegraphics[width=\textwidth]{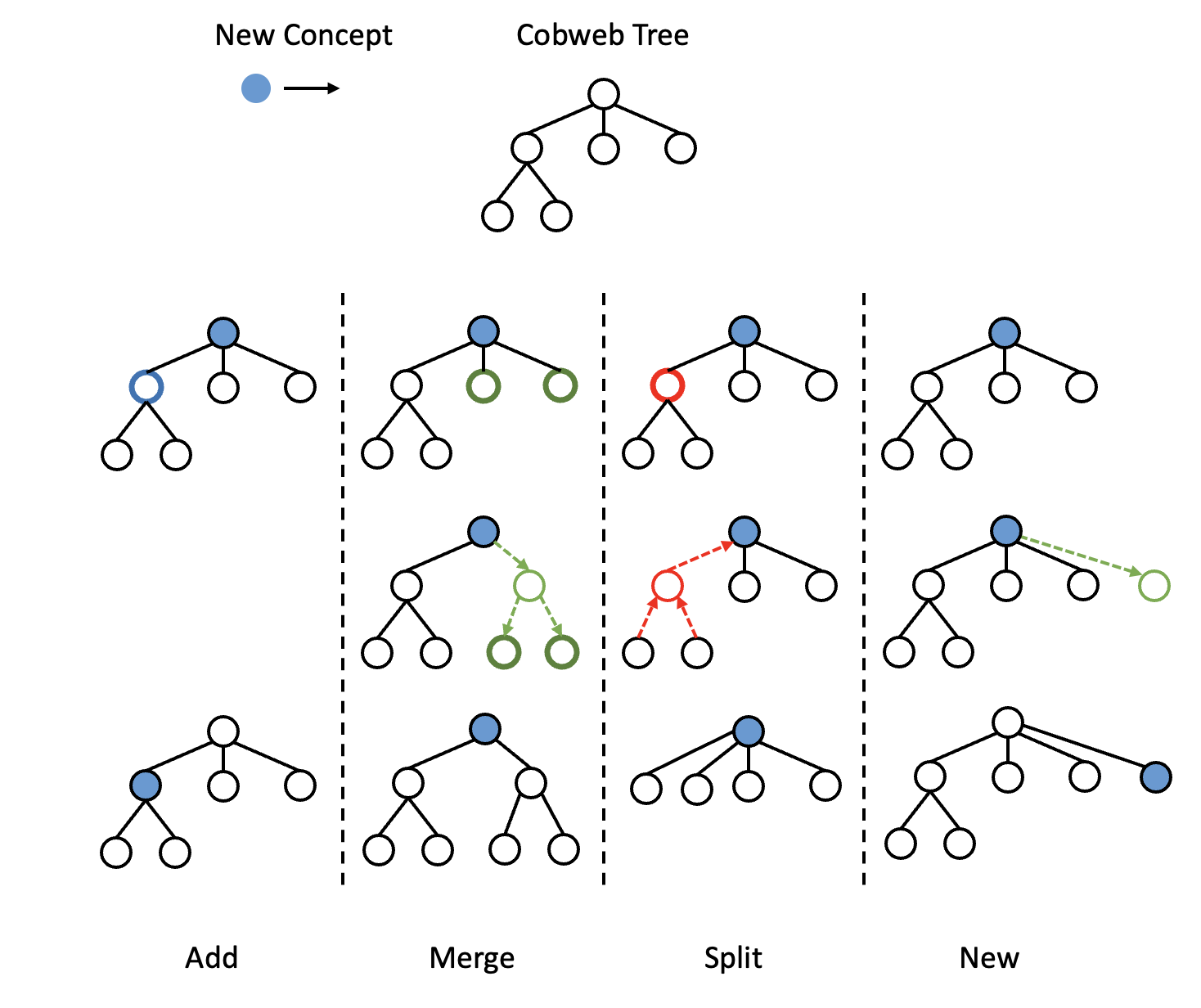}
    \label{fig:cobweb_operations}
    \caption{}
  \end{subfigure}
  \caption{Cobweb's Learning Process. (a) How a new instance is incorporated into the concept hierarchy. (b) The four operations Cobweb applies to update its structure during learning.}
  \label{fig:main}
\end{figure}

Cobweb’s learning mechanism involves classifying each new instance by recursively traversing the hierarchy. Along the selected path, the algorithm updates the count tables of the concept nodes to record the attribute values of the instance (Figure \ref{fig:main}a). At each branching point, Cobweb considers four possible restructuring operations: 
\textit{adding} the instance to the most appropriate child and updating its attribute-value counts, 
\textit{merging} the two most similar children and then reevaluating the available options, 
\textit{splitting} the most similar child and promoting its children to the current level, and \textit{creating} a new child node that initially contains only the new instance (Figure \ref{fig:main}b).

For prediction, Cobweb employs a process similar to its learning procedure, but without updating the concept counts along the path. When presented with a new instance, the algorithm begins at the root of the hierarchy and recursively sorts the instance through the tree. At each branching point, Cobweb decides whether to continue the descent into a child node or to halt at the current concept. Once the traversal ends, the count table of the final concept node is used to estimate the values of any unobserved attributes.
In both its learning and prediction phases, Cobweb relies on a measure known as \textit{category utility} \citep{corter1992explaining} to guide its decisions, 
selecting the operation that yields the highest value. Category utility quantifies the improvement in predictive power offered by a child node compared to its parent. Formally, the measure is defined as:

\begin{equation}
\frac{\sum_{k=1}^{n}P(C_k)\left[\sum_{i}\sum_{j}P(A_i=V_{ij}|C_k)^2-P(A_i=V_{ij})^2\right]}{n}
\end{equation}

Here, $n$ denotes the number of child concepts, $P(C_k)$ is the overall probability of the $k$th child, $P(A_i=V_{ij}|C_k)$ represents the probability that attribute $A_i$ takes value $V_{ij}$ given child $C_k$, 
and $P(A_i=V_{ij})$ is the probability of $A_i$ having value $V_{ij}$ in the parent concept. 
Intuitively, the term $\sum_{i}\sum_{j}P(A_i=V_{ij}|C_k)^2$ captures the expected number of attributes correctly predicted 
within child $C_k$, while $\sum_{i}\sum_{j}P(A_i=V_{ij})^2$ corresponds to the expected number of correct predictions made at the parent level. The category utility score therefore measures the average improvement in predictive accuracy when moving from the parent to its children, with each child weighted by its probability $P(C_k)$. 
To allow comparison across cases with different branching factors, the score is normalized by dividing by $n$, the number of children. 
Although the original Cobweb algorithm supports only nominal attributes, \textit{Cobweb/3} \citep{mckusick1990cobweb} extends the framework to continuous attributes. In this version, each concept models the probability density of continuous attributes using normal distributions, storing the mean and standard deviation for each attribute instead of maintaining nominal count tables.

\subsection{Cobweb/4V - A Novel Version of Cobweb for Image Learning} 
\textit{Cobweb/4V} \citep{barari2024incremental} is an extension of the 
Cobweb framework designed to support continual learning in visual domains. This variant demonstrated two notable capabilities: 
it can learn effectively from limited data and shows strong robustness 
to catastrophic forgetting in sequential visual learning tasks. 
Building on the earlier \textit{Trestle} implementation by \citet{maclellan2016trestle}, this variant introduces several key updates, including an information-theoretic learning measure, a multi-concept prediction strategy, and a tensor-based representation that enables efficient processing of image data.

\subsubsection{Key Updates}
\textit{Learning with Mutual Information:}\\
Earlier Cobweb studies used the \textit{probability-theoretic} category utility \citep{corter1992explaining}, which measures the expected increase in correct attribute predictions given concept membership. 
This formulation has been described as an unsupervised extension of the 
{\it Gini Index} commonly applied in decision tree construction 
\citep{fisher1996iterative}. Cobweb/4V instead employs an \textit{information-theoretic} category utility \citep{corter1992explaining}, linking feature predictability with 
informativeness. The updated measure is defined as:

\begin{equation}
    \frac{\sum_{k=1}^{n}P(C_k)\left[H(A=V) - H(A=V|C_k)\right]}{n}
\end{equation}

where $H(A=V) = \sum_i \sum_j [-P(A_i=V_{ij}) \log(P(A_i=V_{ij})]$ is the entropy over all attribute values in the parent, and $H(A=V|C_k)$ is the entropy for child $k$. This unsupervised extension of information gain, closely related to mutual information, supports greater precision than the probability-theoretic variant. 
By expressing utility in terms of entropy, the approach naturally accommodates different attribute distributions, many of which have closed-form entropy expressions.\\

\noindent \textit{Predicting with a Combination of Concepts:}\\
Traditional Cobweb prediction assigns an instance to a single subordinate-level concept and uses its counts to infer missing attributes \citep{maclellan2016trestle,maclellan2022convolutional,Maclellan2022Efficient}. Other studies have instead used predictions from alternative levels, such as the \textit{basic-level} \citep{fisher1990structure,corter1992explaining}. 
Cobweb/4V introduces a broader approach that combines predictions from multiple concepts in the hierarchy. Given an instance $x$ with unobserved features and a parameter $N_{max}$ (the number of nodes to expand), the system performs a \textit{best-first} search rather than a single-path greedy search. 
At each step, it expands the node $c^*$ on the frontier (the set of candidate nodes awaiting exploration) with the highest score $s(c) = P(c|x)P(x|c)$, known as \textit{collocation} 
\citep{jones1983identifying}, which is the product of cue and category validity. Expanded nodes are collected in $\mathcal{C}^*$, and prediction is made via a softmax-weighted combination of their contributions:

\begin{equation}
    P(X_i=x_i|\mathcal{C}^*) = \sum_{c\in\mathcal{C^*}}P(x_i|c)\frac{\exp\{-s(c)\}}{\sum_{c\in\mathcal{C^*}}\exp\{-s(c)\}}
\end{equation}

Although only $N_{max}$ nodes are expanded, this procedure effectively performs a form of Bayesian model averaging \citep{hinne2020conceptual}.\\

\noindent \textit{A New Tensor Representation:}\\
In Cobweb/4V, instances are represented as tensors of pixel values paired with class labels, rather than lists of attribute-value pairs as in prior versions. For vision tasks, this design allows inputs to be structured as $n$-channel 2D images.
Each node stores the mean and standard deviation of pixel features in a tensor, similar to Cobweb/3’s treatment of continuous attributes, while also maintaining a probability table for class labels. 
Assuming conditional independence among attributes, the uncertainty of a node is computed as the sum of entropies across its attributes. 
Implemented with PyTorch, this tensor-based representation enables faster processing than earlier versions of Cobweb that use attribute-value lists.

\begin{figure}[t!]
    \centering
    \includegraphics[width=0.8\textwidth]{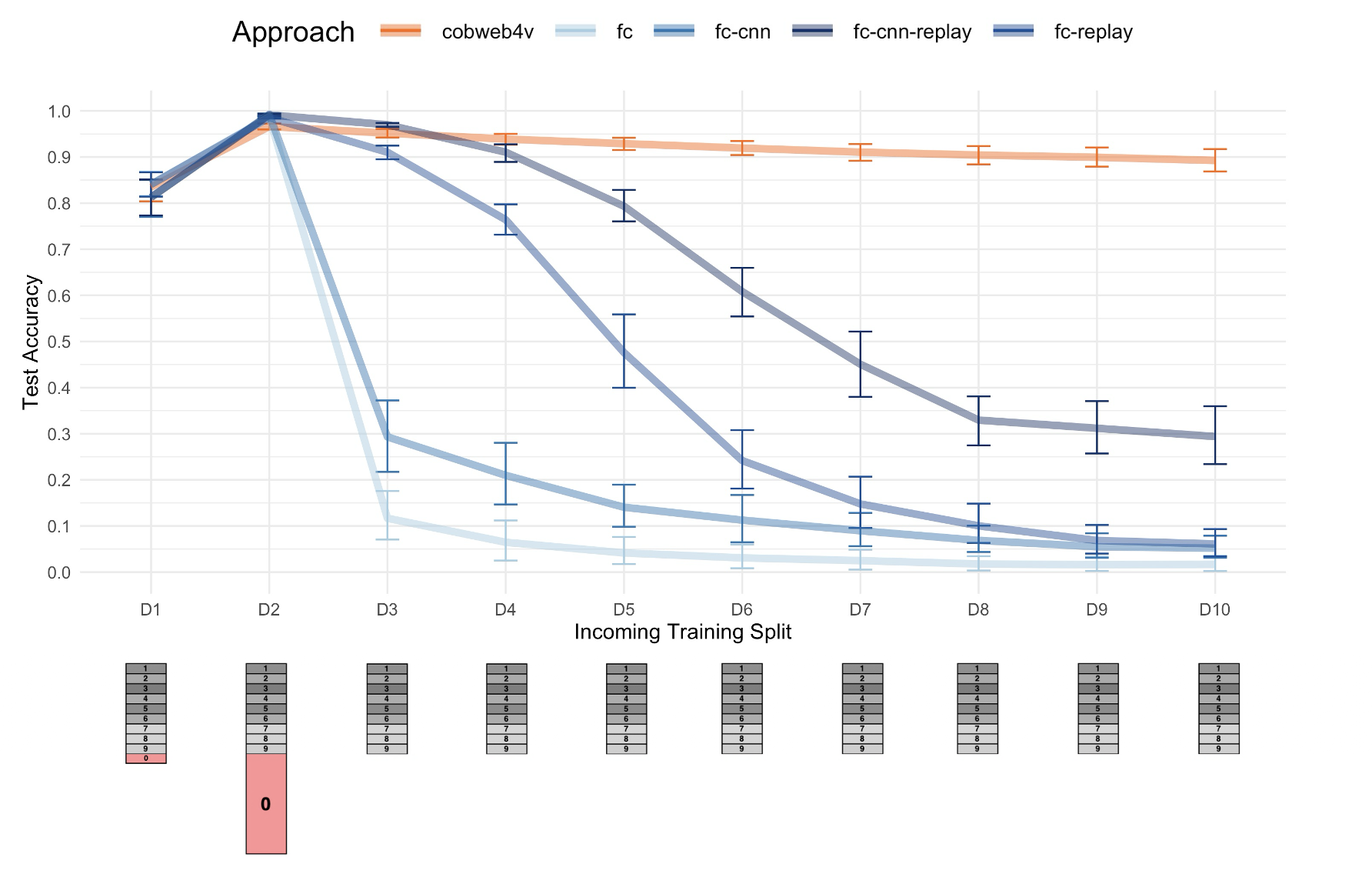}
    \caption{Average test accuracy on the chosen class images from the MNIST test set after each training split (D1--D10). 
    D1 includes a balanced portion of all digits,
    containing 300 images each for digits 0-9. The second split, includes all remaining data
    for the chosen digit, along with an additional 300 images from each of the other non-chosen digits. The remaining data for the non-chosen digits are randomly divided across the remaining 8 splits. The color blocks under the x-axis represent the digit distribution in each split when the chosen digit label is 0.}
    \label{fig:cf_results}
\end{figure}

\subsubsection{Resilient to Catastrophic Forgetting}
Cobweb/4V demonstrates robustness to catastrophic forgetting compared to neural network baselines. The first neural network baseline (fc) employs fully connected layers, and the second baseline (fc-cnn) incorporates additional convolutional neural network (CNN) layers. Figure \ref{fig:cf_results} summarizes the results. Each approach was trained sequentially 
on ten data splits from MNIST dataset and evaluated only on a chosen class after each split, where the chosen class appeared only in the first two training splits and was absent from the rest. 
Neural network baselines without replay exhibited a rapid decline in accuracy, eventually approaching zero. With replay, the network was trained on both the current split and the examples stored in the replay buffer. After each split, 1,000 examples were randomly sampled from the union of the buffer and the split and carried forward for the next training iteration. Even under this replay scheme, the performance of the neural networks decayed steadily as training progressed.
In contrast, Cobweb/4V maintained high accuracy across splits, with only a gradual decline due to feature interference.
These findings indicate that Cobweb/4V preserves prior knowledge effectively, highlighting its resilience to catastrophic forgetting.

\subsection{CobwebNN - A Neural Network Version of Cobweb}
We introduce \textit{CobwebNN}, a neural architecture inspired by, but distinct from neural taxonomic networks \citep{wang2025taxonomic}. Neural taxonomic nets organize concepts hierarchically, using gating functions for branching, classifiers for label prediction, and mechanisms such as temperature-controlled gating, stochastic exploration with Gumbel noise, and regularization for balanced splits. These techniques make them effective for differentiable concept hierarchies.
CobwebNN adapts this hierarchical idea but is designed to approximate the behavior of the Cobweb/4V framework. Unlike taxonomic nets that rely on gating and linear classifiers, CobwebNN represents concepts through reconstruction-based prototypes coupled with class distributions. This design enables controlled comparisons with Cobweb/4V to examine the role of structural non-gradient-based learning in mitigating catastrophic forgetting.
%
% Each input instance x is matched to concept nodes according to reconstruction quality. Each node maintains (1) a prototype for reconstruction, (2) a prior weight for baseline likelihood, and (3) label information for classification. 

%
In CobwebNN, each input instance $x$ is matched to concept nodes based on how well it aligns with their prototypes. To support this process, each node maintains three components:
(1) Prototype for reconstruction: 
% a vector representation that summarizes the typical input associated with a concept $c$. The match between an input $x$ and this prototype is evaluated using a Gaussian likelihood with unit variance, $p(x \mid c)$. This serves both as a reconstruction of the concept and as a probabilistic measure of similarity. The variance is fixed to one so that all concepts are compared on the same scale, ensuring that likelihoods depend only on how close an input is to a prototype. 
a vector $\mu_c$ that summarizes the typical input associated with a concept $c$. The match between an input $x$ and this prototype is evaluated using a Gaussian likelihood with unit variance, $p(x \mid c) = \mathcal{N}(x; \mu_c, I)$. This likelihood both reflects reconstruction quality, since $\mu_c$ serves as the representative input for the concept, and provides a probabilistic measure of similarity, with higher values assigned to inputs closer to the prototype. The variance is fixed to one so that all concepts are compared on the same scale, ensuring that likelihoods depend only on how close an input is to a prototype. (2) Prior weight for baseline likelihood: a learnable bias term that specifies the baseline probability of selecting a child concept given its parent, denoted $p(c \mid c_{\text{parent}})$. This ensures that every branch retains some probability mass even before considering the input. Intuitively, it is the parent’s “default preference” for its children, later modulated by the data. (3) Label information for classification: a set of learnable parameters that define $p(y \mid c)$, the probability of predicting each label when concept $c$ is chosen.

Path selection is made differentiable through the Gumbel–Softmax trick \citep{jang2017categoricalreparameterizationgumbelsoftmax}, which approximates categorical sampling in a continuous and differentiable way. It adds Gumbel noise to the logits, followed by a temperature-controlled softmax. The temperature parameter $\tau$ adjusts how close the output is to a one-hot vector: lower $\tau$ yields sharper, more discrete selections, while higher $\tau$ produces smoother probabilities. This allows gradient flow through otherwise discrete branching.
Formally, the path probability at layer $L$ is defined recursively as:
\begin{equation}
p^{L}(c \mid x) \thicksim p(x \mid c) \cdot p^{L-1}(c_{\text{parent}} \mid x) \cdot p(c \mid c_{\text{parent}})
\end{equation}
with $p^{0}(\text{root} \mid x) = 1$. 
Here $x$ is the input instance, $y$ is the label, $c$ is the current concept node, $c_{parent}$ is the parent node of concept $c$, and $C_L$ is the set of nodes in level $L$ of the hierarchy.

Label prediction marginalizes over all leaves:
\begin{equation}
p(y \mid x) = \sum_{c \in \mathcal{C}_L} p^{L}(c \mid x) \cdot p(y \mid c)
\end{equation}
where $\mathcal{C}_L$ denotes the set of concepts in the final level (leaves), and $p(y \mid c)$ is a softmax over learnable logits.
Two inference modes are supported: 1. Sparse mode; a single leaf is sampled using Gumbel-Softmax, and prediction is drawn from its label distribution. 2. Dense mode; all leaves contribute, weighted by their path probabilities.

Training follows the same principle, except path probabilities incorporate both features and labels:
\begin{equation}
p^{L}(c \mid x,y) \thicksim p(x,y \mid c) \cdot p^{L-1}(c_{\text{parent}} \mid x,y) \cdot p(c \mid c_{\text{parent}})
\end{equation}
We factor the joint distribution as $p(x,y \mid c) = p(x \mid c) \cdot p(y \mid c)$, assuming conditional independence of the input features and labels given the concept. In this view, $p(x \mid c)$ captures how well the instance matches the concept’s prototype, while $p(y \mid c)$ captures the label distribution associated with that concept.
The objective combines reconstruction and classification, with loss:
\begin{equation}
\mathcal{L} = - \frac{1}{N} \sum_{n=1}^N \sum_{c \in \mathcal{C}_L} p^L(c \mid x_n, y_n) \cdot \Big( \log p(x_n \mid c) + \log p(y_n \mid c) \Big)
\end{equation}
Thus, the loss updates each concept in proportion to both its ability to reconstruct the input and its ability to predict the correct label.
In dense training, all nodes are updated proportionally to their path probabilities. In sparse training, only nodes along one sampled path are updated, mirroring inference and reducing interference from unrelated concepts.

The results presented in prior work showed that Cobweb/4V is able to learn visual concepts in a continual setting while exhibiting strong robustness to catastrophic forgetting, even when compared against competitive neural network baselines that incorporate replay strategies. 
While these findings established Cobweb/4V as a promising framework for continual learning, they also raised an important question: 
\textit{what underlying mechanisms enable this resilience?} 
Addressing this question is essential not only for understanding the principles behind Cobweb/4V, but also for identifying features that could inform the design of future continual learning systems. Motivated by this goal, the next section introduces three hypotheses that seek to explain Cobweb/4V’s robustness to forgetting, each grounded in distinct characteristics of the framework’s structure and learning dynamics.

\section{Forgetting Hypotheses}
Although Cobweb/4V has demonstrated strong resilience to catastrophic forgetting in continual visual learning tasks, the reasons for this robustness remain unclear. To investigate, we propose three complementary hypotheses that each target a different potential contributing factor. The first examines the role of Cobweb’s adaptive hierarchical structure, the second considers its use of sparse and localized updates during learning, and the third focuses on its information-theoretic learning mechanism as an alternative to gradient-based optimization. By testing these hypotheses across multiple datasets and controlled variations of the framework, we aim to identify the key properties that enable Cobweb/4V to mitigate catastrophic forgetting and assess their broader relevance to continual learning models.

\subsection{Adaptive Structure}
A well-known challenge in continual learning is the limited capacity of learning models. Systems with fixed or constrained structures often struggle to integrate new knowledge without overwriting previously learned information, as interference arises when old and new data compete for the same representational resources \citep{french1999catastrophic, parisi2019continual}. This trade-off between plasticity and stability is a central factor in catastrophic forgetting. 
Cobweb’s adaptive structure provides a potential means of addressing this challenge. During learning, Cobweb dynamically reorganizes its hierarchy by creating, merging, or splitting nodes in response to new data. This allows the model to allocate representational capacity where it is most needed and to adjust its concept hierarchy as distributions shift. In principle, such 
structural flexibility could mitigate forgetting by reducing interference between old and new knowledge. 
Based on these observations, we hypothesize that Cobweb’s adaptive structure plays an important role in its robustness to catastrophic forgetting. To evaluate this hypothesis, we design experiments that compare the original adaptive Cobweb to a fixed-structure variant, allowing us to isolate the effect of structural adaptivity on continual learning performance.

\subsection{Sparse Updates}
Another major contributor to catastrophic forgetting is how models update their internal representations when new data arrive. In neural networks, learning typically occurs through backpropagation, where all parameters are adjusted at once for each new batch. This dense updating can cause interference: changes made for new information may overwrite weights that were essential for earlier knowledge, leading to forgetting over time \citep{goodfellow2013empirical}.
Cobweb, in contrast, relies on sparse and selective updates. When an input is categorized, only a single path, or a small localized part of the hierarchy, is updated, leaving unrelated knowledge intact. This localized adaptation reduces interference and mirrors the sparse activation patterns observed in biological learning, which help preserve memory by limiting overlap across tasks \citep{olshausen1996emergence, masse2018alleviating}.
We therefore hypothesize that Cobweb’s reliance on sparse, localized updates plays a key role in its robustness to catastrophic forgetting. To test this, we compare the standard sparse-update process to a dense-update alternative, where a broader portion of the hierarchy is modified during learning, allowing us to isolate the effect of update sparsity on interference.

\subsection{Information-Theoretic Learning Mechanism}
A further potential explanation for Cobweb’s robustness to catastrophic forgetting lies in its learning mechanism. Most neural networks rely on backpropagation, which performs iterative gradient-based updates over mini batches of data. While effective for task-specific optimization, this process introduces a recency bias: parameter estimates are influenced more strongly by the most recent data, leading to a gradual disruption of information about earlier experiences \citep{goodfellow2013empirical}. Since old data are not revisited in continual learning settings, gradient descent updates often approximate a moving average rather than a true posterior, which increases the risk of forgetting. 
In contrast, Cobweb employs a closed-form, information-theoretic learning mechanism that leverages sufficiency statistics under the assumption of normal distributions. Each concept tracks the number of instances seen, as well as the mean and variance of their feature values. These statistics are sufficient in the statistical sense: they retain all the information the data provide about the distribution’s parameters. As a result, Cobweb can update its concept representations incrementally with each new instance while maintaining unbiased estimates of the mean and variance across all data observed, without the need to revisit earlier examples. This approach effectively avoids the recency bias inherent in stochastic gradient descent, allowing the system to preserve prior knowledge more faithfully. 
Based on these observations, we hypothesize that Cobweb’s information-theoretic learning mechanism, supported by the use of sufficiency statistics, plays a critical role in its robustness to catastrophic forgetting. To evaluate this hypothesis, we design experiments that directly compare Cobweb’s closed-form updates with gradient-based optimization methods, allowing us to assess the extent to which the choice of learning mechanism contributes to its stability in continual learning.

\section{Experiments}
Our experiments aim to investigate the mechanisms underlying Cobweb/4V’s robustness to catastrophic forgetting. To ensure the results are not tied to a single benchmark, we evaluate the model on a range of image datasets that differ in content and complexity. We introduce three main sets of experiments that each test one of the proposed hypotheses.

\subsection{Datasets}
To evaluate generalizability, we test on four widely used image datasets that differ in content and complexity: handwritten digits, clothing items, natural images, and medical imagery.
\begin{itemize}

\item MNIST \citep{lecun1998mnist} contains 70,000 grayscale images of handwritten digits of size $28 \times 28$. It is a long-standing benchmark in continual learning due to its simplicity and balanced classes.

\item Fashion-MNIST \citep{xiao2017fashion} has the same format as MNIST but depicts clothing items (e.g., shirts, trousers, sneakers, bags). With 70,000 grayscale images at $28 \times 28$, it poses a more visually challenging task than handwritten digits.

\item CIFAR-10 \citep{krizhevsky2009learning} consists of 60,000 color images of size $32 \times 32$ from ten classes of natural objects such as animals, vehicles, and household items. Its greater visual diversity makes it a stronger test of robustness compared to grayscale benchmarks.

\item MedMNIST (OrganA subset) \citep{yang2023medmnist} provides multi-class abdominal organ images derived from medical scans. Designed as a lightweight medical benchmark, it introduces more realistic scenarios where resilience to forgetting is critical. The OrganA subset contains 58,850 grayscale images at $28 \times 28$ resolution, spanning eleven abdominal organ classes such as liver, spleen, kidney, and stomach.
    
\end{itemize}

\noindent Together, these datasets form a diverse testbed for assessing both the stability and adaptability of Cobweb/4V and its neural variants in continual learning.

\subsection{Training Splits}
Following the protocol from our previous work \citep{barari2024incremental}, we partition each dataset into ten splits (D1–D10). The first split (D1) contains a balanced sample of all classes, with 300 images per class. The second split (D2) consists of all remaining data for the chosen class, together with an additional 300 images from each of the other non-chosen classes. The remaining data from the non-chosen classes are then randomly and evenly distributed across the last eight splits (D3–D10).

\begin{figure}[t!]
    \centering
    \includegraphics[width=\textwidth]{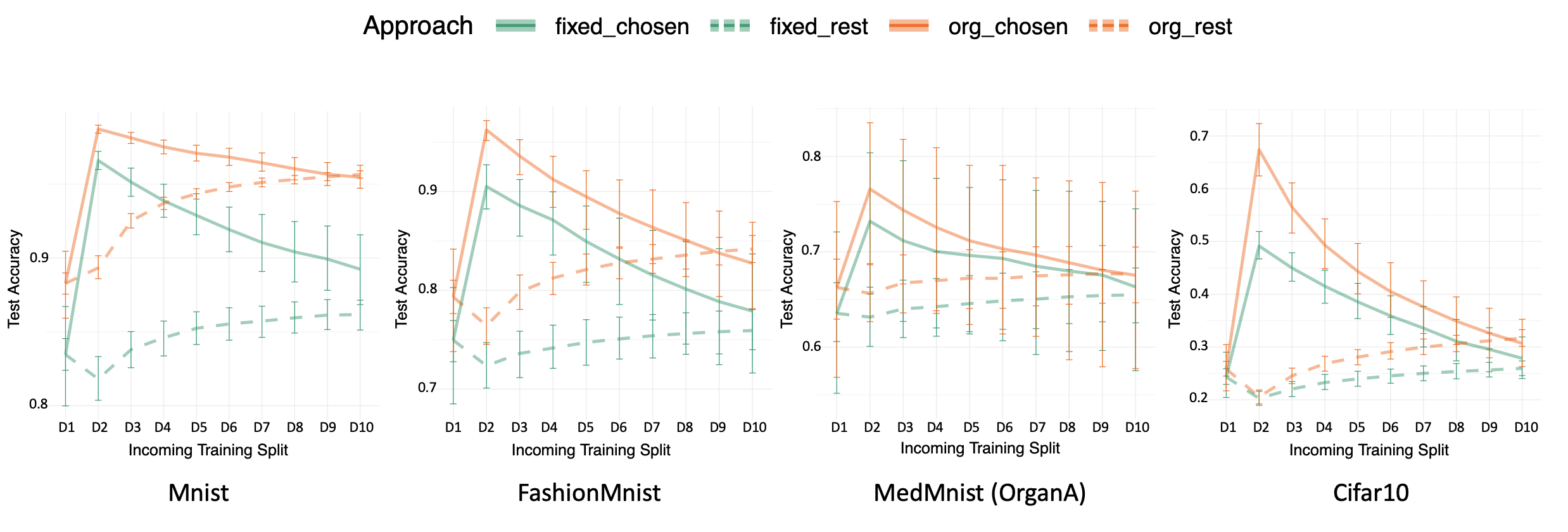}
    \caption{Average accuracy of (Fixed vs. Adaptive)-structure Cobweb/4V on Chosen and Non-chosen classes across datasets, after each training split (D1--D10). 
    Solid lines represent accuracy on the chosen class; dashed lines represent average accuracy on non-chosen classes. \textit{fixed\_} refers to the fixed-structure Cobweb/4V and \textit{org\_} refers to the original Cobweb/4V with adaptive structure.}
    \label{fig:exp1}
\end{figure}

\subsection{Experiment 1 - Adaptive vs. Fixed Structure}
\subsubsection{Method}
This experiment evaluates the first hypothesis: that Cobweb’s adaptive structure contributes to its robustness to catastrophic forgetting. Cobweb/4V normally employs dynamic restructuring operations, including creating, merging, and splitting nodes, which allow the hierarchy to expand and reorganize as new data are introduced. 
To test the role of structural adaptivity, we compare the standard adaptive version of Cobweb/4V with a fixed-structure variant. In the fixed version, the depth and branching factor of the tree are predetermined, and merge and split operations are disabled. This design removes the system’s ability to dynamically reorganize its structure while maintaining all other aspects of the learning process. Both variants are trained under the continual learning protocol described earlier, using sequential data splits across multiple dataset. Performance is evaluated on both the chosen class and the non-chosen classes after each split, allowing us to assess differences in knowledge retention and the ability to incorporate new information. 

\subsubsection{Results}
Figure \ref{fig:exp1} shows that fixing the structure consistently reduced accuracy compared to adaptive Cobweb/4V, underscoring the importance of reorganization for both stability and plasticity.
Even so, the fixed-structure model maintained relatively stable performance across training. It did not exhibit sharp drops in accuracy, and forgetting remained gradual. Earlier knowledge was largely preserved, suggesting that factors beyond structural adaptivity also contribute to Cobweb/4V’s robustness. 

\subsubsection{Discussion}
These findings show that Cobweb/4V’s adaptive structure strengthens both memory stability and learning plasticity. By allowing nodes to be created, merged, and split, the model integrates new concepts while preserving prior knowledge, leading to consistently higher performance than the fixed-structure variant on both chosen and non-chosen classes.
At the same time, the fixed-structure model still demonstrated notable robustness to forgetting. Its accuracy declined gradually rather than collapsing sharply, suggesting that structural adaptivity is important but not the sole driver of Cobweb/4V’s stability. Importantly, robustness cannot be explained simply by Cobweb's instance-based design. Even when leaves no longer corresponded to individual training examples, the model retained stable performance by relying on intermediate concepts rather than memorized exemplars.
Taken together, these results support the view, consistent with cognitive science, that learning involves updating and reorganizing internal structures to accommodate new information while maintaining continuity of past knowledge.

\subsection{Experiment 2 - Sparse vs. Dense Updates}
\subsubsection{Method}
This experiment tests the hypothesis that Cobweb’s resistance to catastrophic forgetting arises from its sparse and selective updates. In Cobweb/4V, a new instance updates only a single path or a small subset of nodes in the hierarchy, unlike neural networks where backpropagation adjusts all parameters at once. Such localized updates are expected to reduce interference between old and new knowledge.
To evaluate this idea, we used CobwebNN, a neural architecture that mimics Cobweb while allowing explicit control over update sparsity via the Gumbel-Softmax trick \citep{jang2017categoricalreparameterizationgumbelsoftmax, wang2025taxonomic}. By adjusting the temperature parameter $\tau$ and sampling mode, CobwebNN can be run in either sparse-update mode (one path) or dense-update mode (multiple paths).
Both variants were trained under the continual learning protocol described earlier. Performance was then measured on the chosen class to track forgetting, and on non-chosen classes to test generalization to new data across splits.

\begin{figure}[t!]
    \centering
    \includegraphics[width=\textwidth]{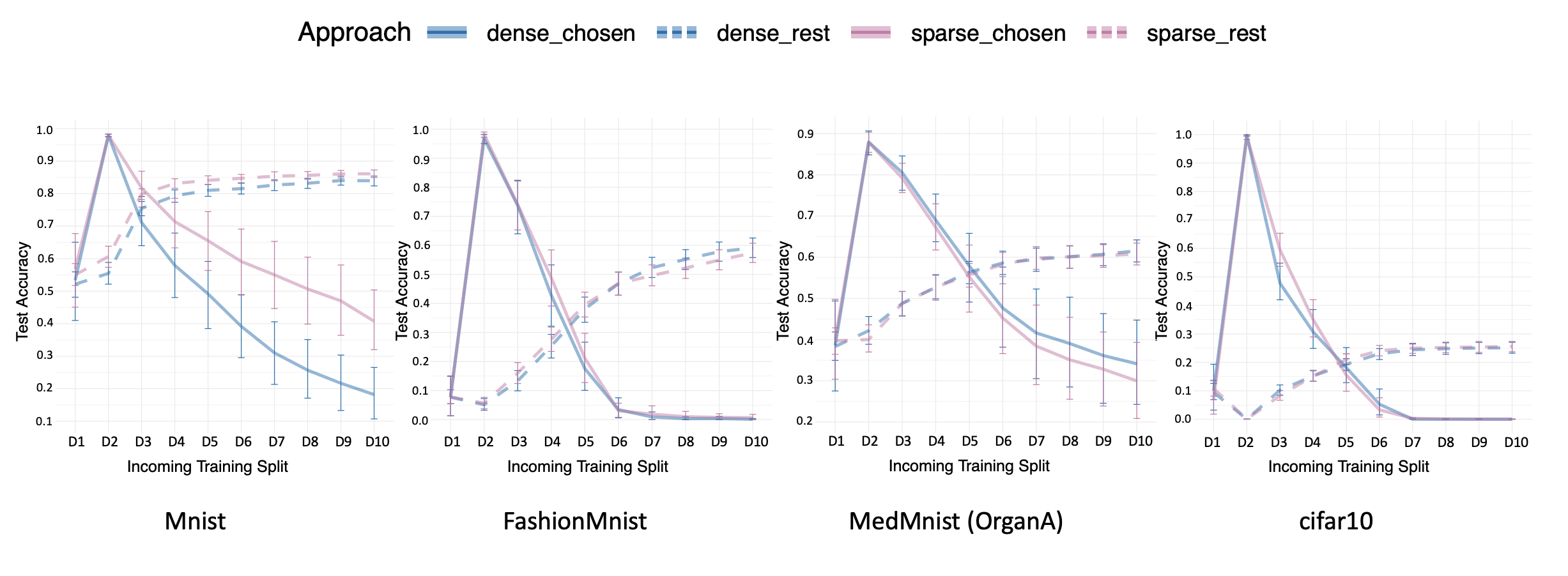}
    \caption{Average accuracy of (Sparse vs. Dense)-update configurations of CobwebNN on Chosen and Non-chosen classes across datasets. Each subplot shows the test accuracy after each training split. Solid lines represent accuracy on the chosen class; dashed lines represent average accuracy on non-chosen classes.}
    \label{fig:exp2}
\end{figure}

\subsubsection{Results}
Figure \ref{fig:exp2} compares sparse- and dense-update variants of CobwebNN on both chosen and non-chosen classes. Results show no substantial accuracy difference between the two modes. In both cases, chosen-class performance declined gradually across training splits, reflecting forgetting, while non-chosen class performance followed similar trends. These findings suggest that, in this implementation, update sparsity did not measurably affect either memory retention or learning of new information.

\subsubsection{Discussion}
This experiment tested whether sparse updates, as in Cobweb/4V, reduce interference and improve retention in a neural network setting. The results are inconclusive: no clear performance gap emerged between sparse and dense updates. While sparsity may provide benefits under certain conditions, in CobwebNN other factors, such as the learning mechanism and absence of hierarchical restructuring, likely dominate forgetting. Thus, we find no definitive evidence for the effect of sparsity on continual learning in neural networks. Further work that isolates sparsity from such confounding factors will be needed to clarify its role.

\subsection{Experiment 3 - Information-Theoretic vs. Backpropagation}
\subsubsection{Method}
The third experiment examines whether Cobweb/4V’s resilience to catastrophic forgetting stems from its information-theoretic learning mechanism rather than the gradient-based optimization used in neural networks. To isolate this effect, we compare fixed-structure Cobweb/4V with the sparse-update variant of CobwebNN, ensuring both have similar structural sparsity.
In Cobweb/4V, each concept node is modeled as a multivariate normal distribution with diagonal covariance. The model maintains sufficiency statistics (count, mean, variance), which allow incremental Bayesian updates without revisiting past data. For example, given $N$ prior observations (the number of examples already seen for the concept) the mean and variance can be updated with a new input $x_{\text{new}}$ as:
\begin{equation}
\mu_{\text{new}} = \mu_{\text{old}} + \frac{1}{N+1}(x_{\text{new}} - \mu_{\text{old}})
\end{equation}

\begin{equation}
\sigma^2_{\text{new}} = \sigma^2_{\text{old}} + \frac{1}{N+1}\big((x_{\text{new}} - \mu_{\text{old}})(x_{\text{new}} - \mu_{\text{new}}) - \sigma^2_{\text{old}}\big)
\end{equation}
As data accumulate, the effect of each new input diminishes, reflecting the growing evidence base. This process yields unbiased posterior estimates equivalent to batch computation, while avoiding storage or replay of past examples.
These updates are mathematically equivalent to a gradient-based update (Equation 10) with a learning rate of $1/(1+N)$. In contrast, sparse CobwebNN relies on gradient descent and backpropagation. Even with structural sparsity, its updates resemble an exponential moving average, where each input contributes a fixed proportion regardless of prior experience:
\begin{equation}
\theta_{t+1} = \theta_t - \alpha \nabla_{\theta} L(x_t).
\end{equation}
Here, $\theta$ denotes the learnable parameters of the network, $t$ indexes the training step, and $x_t$ is the input processed at step $t$. This uniform weighting makes the network more prone to interference from recent inputs and thus more vulnerable to forgetting. Both models were evaluated under the same continual learning protocol, with performance on the chosen and non-chosen classes tracked after each training split. 

\begin{figure}[t!]
    \centering
    \includegraphics[width=\textwidth]{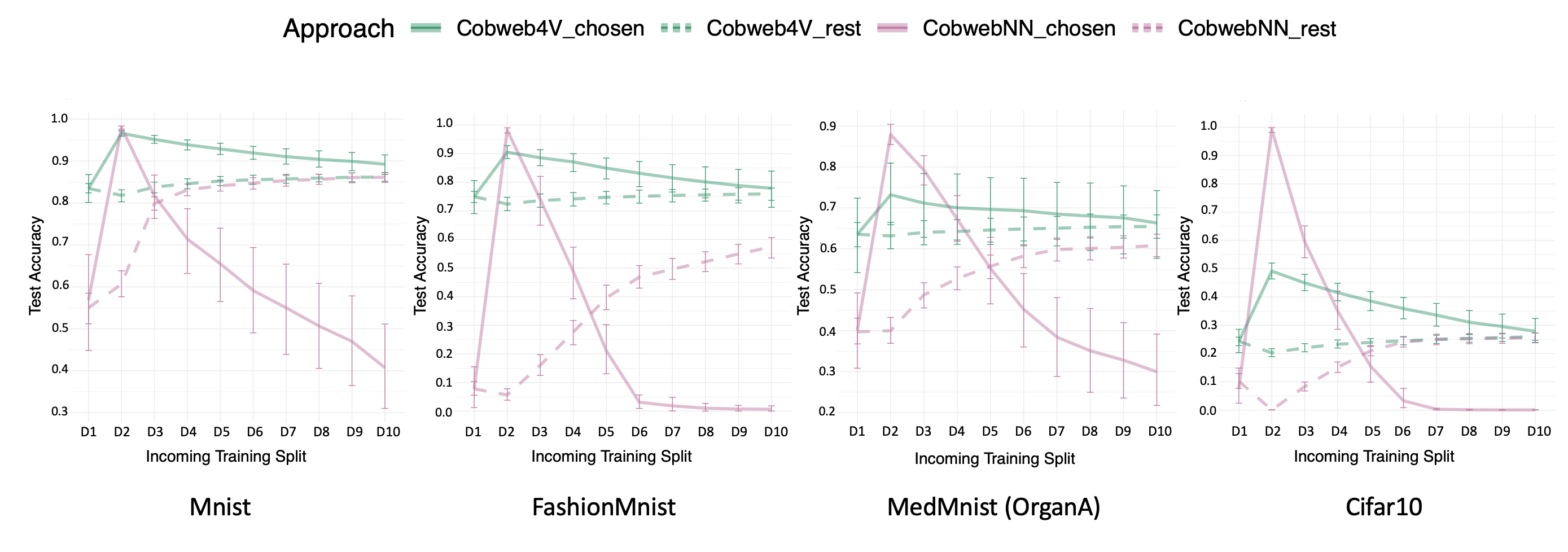}
    \caption{Average accuracy of fixed Cobweb4V vs. sparse CobwebNN on Chosen and Non-chosen classes across datasets. Each subplot shows the test accuracy after each training split. Solid lines represent accuracy on the chosen class; dashed lines represent average accuracy on non-chosen classes.}
    \label{fig:exp3}
\end{figure}

\subsubsection{Results}
Figure~\ref{fig:exp3} compares fixed Cobweb/4V with sparse CobwebNN across datasets. For the chosen class (solid lines), CobwebNN showed a sharp accuracy decline over successive splits, a clear sign of catastrophic forgetting. In contrast, Cobweb/4V maintained stable accuracy, demonstrating stronger resistance to forgetting.
For the non-chosen classes (dashed lines), both models improved as new tasks were introduced, but Cobweb/4V consistently outperformed CobwebNN, with the gap narrowing as training progressed.
Overall, these results indicate that Cobweb/4V preserves earlier knowledge more effectively while still supporting new learning, underscoring the stability-plasticity balance provided by its information-theoretic learning mechanism.  

\subsubsection{Discussion}
These findings suggest that Cobweb/4V’s resistance to catastrophic forgetting stems largely from its information-theoretic learning process. By maintaining sufficiency statistics and updating parameters in closed form, the model incorporates new data while preserving essential information about past inputs, eliminating the need to revisit earlier examples. In contrast, CobwebNN relies on parameter updates similar to a moving average, which gradually overwrite older experiences with recent ones.
Even when structural configurations are matched, the two models diverge in performance, showing that the learning mechanism itself plays a key role in knowledge retention. Cobweb/4V’s probability-based updates allow it to integrate new information in a principled and stable manner, highlighting that algorithmic design, beyond structural adaptivity or sparsity, is central to mitigating catastrophic forgetting in continual learning.

\section{Conclusion and Future Work}
This study examined the factors behind Cobweb/4V’s resilience to catastrophic forgetting in continual learning. Three hypotheses were tested: (1) structural reorganization enhances stability, (2) sparse and selective updates reduce interference, and (3) an information-theoretic learning mechanism supports memory retention.
Results show that while restructuring improves flexibility, it is not the main driver of stability. Strongest support emerges for the third hypothesis: Cobweb/4V’s use of sufficiency statistics enables accurate, incremental updates without revisiting past data, substantially reducing forgetting.
Overall, Cobweb/4V’s stability appears to arise from interacting factors, with its information-theoretic learning as the key contributor. These findings highlight the value of concept-based, probabilistic models as an alternative to gradient-based methods for continual learning.
Future work should investigate how neural models can adopt Cobweb’s adaptive learning dynamics by adjusting their update rules to scale with accumulated experience. Instead of using a fixed learning rate, such models could gradually reduce update magnitudes as evidence grows, similar to Cobweb’s incremental learning behavior. This could combine the representational power of neural networks with the statistical stability of concept-based learning, advancing the development of continual learning models that maintain prior knowledge while adapting to new information.

%\vspace{0.5in}

{\parindent -10pt\leftskip 10pt\noindent
\bibliographystyle{cogsysapa}
\bibliography{format}

}

% Leave a blank line before the closing brace to ensure the final 
% reference has the proper indentation. 

\end{document}